\documentclass[10pt,twocolumn,letterpaper]{article}

\usepackage{wacv}
\usepackage{times}
\usepackage{epsfig}
\usepackage{graphicx}
\usepackage{amsmath}
\usepackage{amssymb}
\usepackage{multirow}
\usepackage{subcaption}
\usepackage{multicol}
\usepackage{dblfloatfix}
\usepackage[normalem]{ulem}
\def\wacvPaperID{1267} 

\wacvfinalcopy 
\ifwacvfinal
\fi

\ifwacvfinal
\usepackage[breaklinks=true,bookmarks=false]{hyperref}
\else
\usepackage[pagebackref=true,breaklinks=true,colorlinks,bookmarks=false]{hyperref}
\fi

\ifwacvfinal
\else
\fi

\begin{document}

\title{GlocalNet: Class-aware Long-term Human Motion Synthesis}

\author{Neeraj Battan \thanks{Indicates equal contribution}, Yudhik Agrawal \footnotemark[1],  Sai Soorya Rao, Aman Goel, and Avinash Sharma\\
International Institute of Information Technology, Hyderabad\\
{\tt\small \{neeraj.battan, yudhik.agrawal\}@research.iiit.ac.in,}\\
{\tt\small \{sai.soorya, aman.goel\}@students.iiit.ac.in, asharma@iiit.ac.in}
}
\maketitle
\begin{abstract}
Synthesis of long-term human motion skeleton sequences is essential to aid human-centric video generation \cite{chan2019everybody} with potential applications in Augmented Reality, 3D character animations, pedestrian trajectory prediction, etc.
Long-term human motion synthesis is a challenging task due to multiple factors like, long-term temporal dependencies among poses, cyclic repetition across poses, bi-directional and multi-scale dependencies among poses, variable speed of actions, and a large as well as partially overlapping space of temporal pose variations across multiple class/types of human activities. This paper aims to address these challenges to synthesize a long-term ($> 6000$ ms) human motion trajectory across a large variety of human activity classes ($>50$).
We propose a two-stage activity generation method to achieve this goal, where the first stage deals with learning the long-term global pose dependencies in activity sequences by learning to synthesize a sparse motion trajectory while the second stage addresses the generation of dense motion trajectories taking the output of the first stage. We demonstrate the superiority of the proposed method over SOTA methods using various quantitative evaluation metrics on publicly available datasets.
\end{abstract}

\section{Introduction}

\label{sec:intro}

Skeleton sequences are traditionally used for human activity/action representation \& analysis~\cite{si2019attention}. Recently, human motion synthesis 
\cite{barsoum2018hp,butepage2017deep,cai2018deep,fragkiadaki2015recurrent,li2017auto,martinez2017human} is gaining ground as it is widely used to aid human-centric video generation \cite{chan2019everybody} with potential applications in Augmented Reality, 3D character animations, pedestrian trajectory prediction, etc.

Human motion synthesis is a challenging task due to multiple factors like long-term temporal dependencies among poses, cyclic repetition across poses, bi-directional and multi-scale dependencies among poses, variable speed of actions, and a large as well as partially overlapping space of temporal pose variations across multiple class/types of human activities.  
 %
%

\begin{figure}[ht!]
\centering
\begin{subfigure}[b]{0.46\textwidth}
  \includegraphics[width=1\linewidth,height=3.5cm]{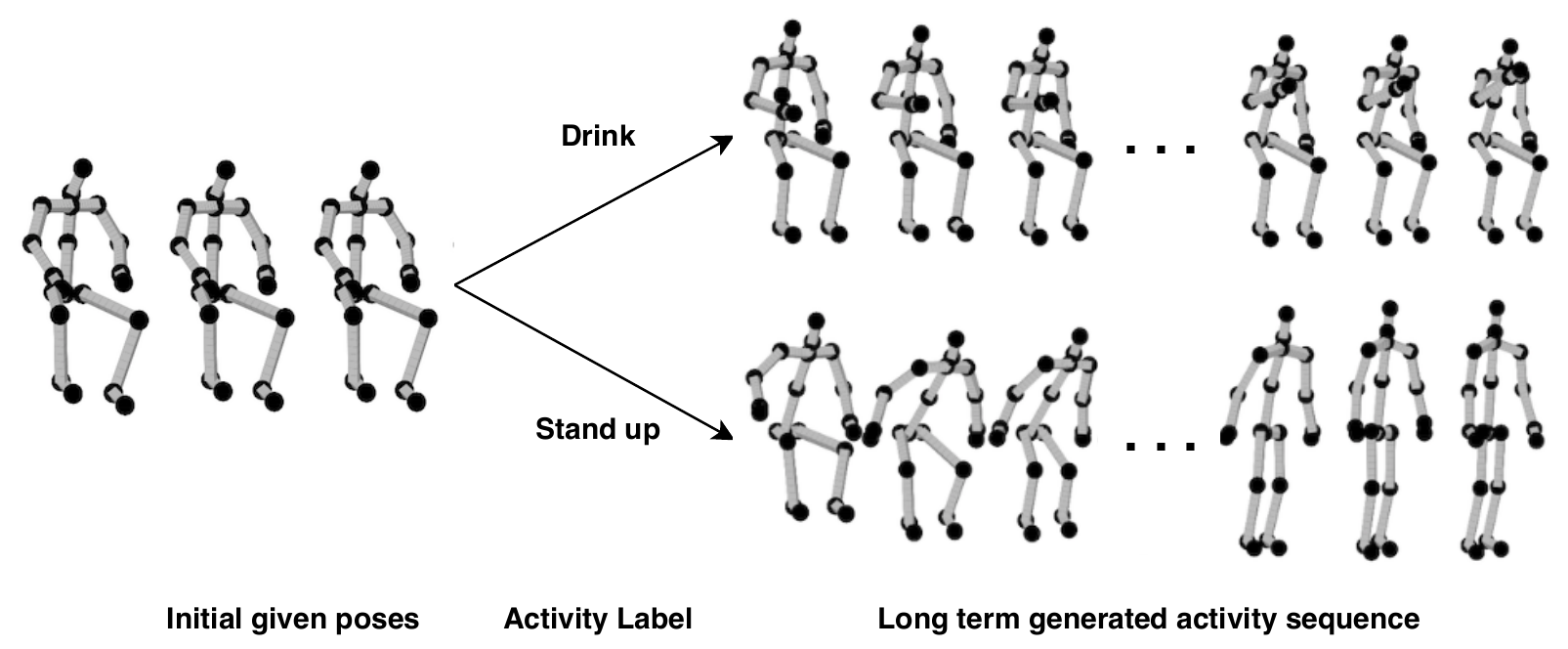}
  \caption{}
  \label{fig:motivation_fig_initial_pose} 
\end{subfigure}

\begin{subfigure}[b]{0.46\textwidth}
  \includegraphics[width=1\linewidth]{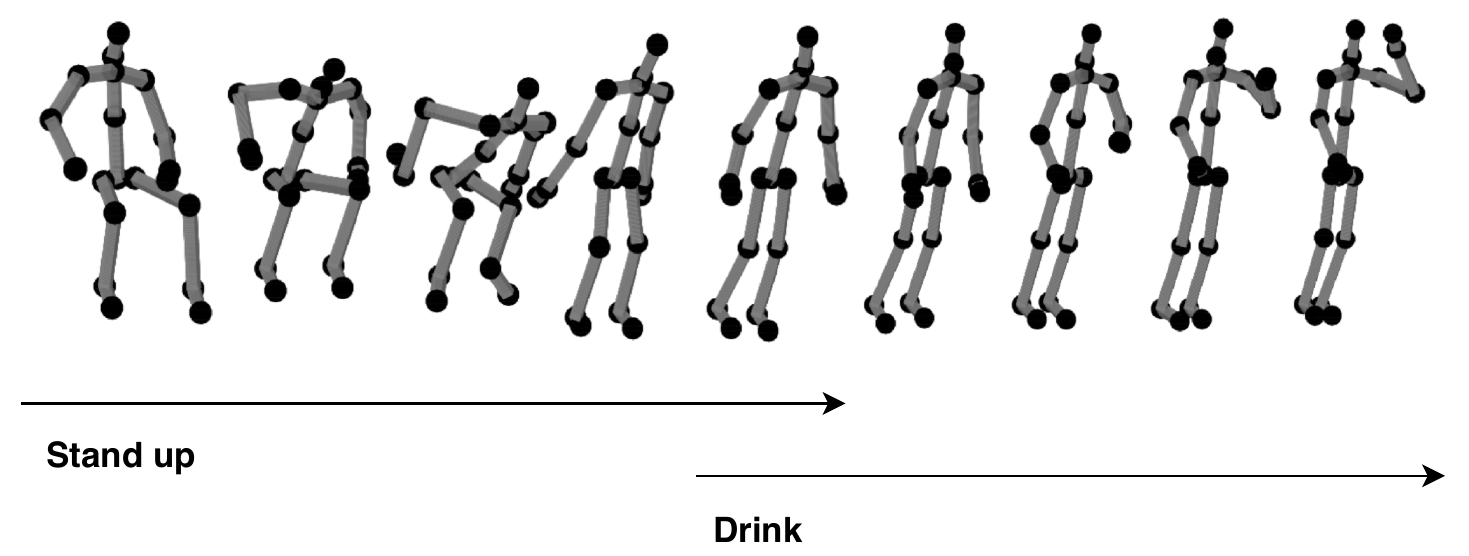}
  \caption{}
  \label{fig:motivation_fig_seq_example}
\end{subfigure}

\caption[]{Motivation: a) Using the same set of sparse initial poses, our method can generate differents type of activities based on the input class label. The figure depicts two such activities - Drinking and Standing up that were synthesized from the same set of initial poses. b) Our method is also capable of transitioning across actions. The figure demonstrates the transition from Standing Up to Drinking activity.
\label{fig:motivation_fig}
}
\end{figure}
Existing methods for human motion synthesis \cite{barsoum2018hp,butepage2017deep,fragkiadaki2015recurrent,ghosh2017learning,holden2016deep,li2017auto} primarily uses auto-regressive models such as LSTM \cite{hochreiter1997long}, GRU \cite{bahdanau2014neural} and Seq2Seq \cite{sutskever2014sequence} which aim to predict a temporally short-duration motion trajectories (of near future) given a set of few initial poses (or sometime referred as frames). 
However, these models do not generalize well while generating long-duration motion trajectories across multiple activity classes due to following inherent limitations. First, typically these auto-regressive models are fed with temporally redundant poses and thus their Markovian dependency assumption fails to exploit the long-duration dependencies among poses. 
\begin{figure*}[ht!]
\begin{center}
\includegraphics[width=0.9\linewidth,height=8.5cm]{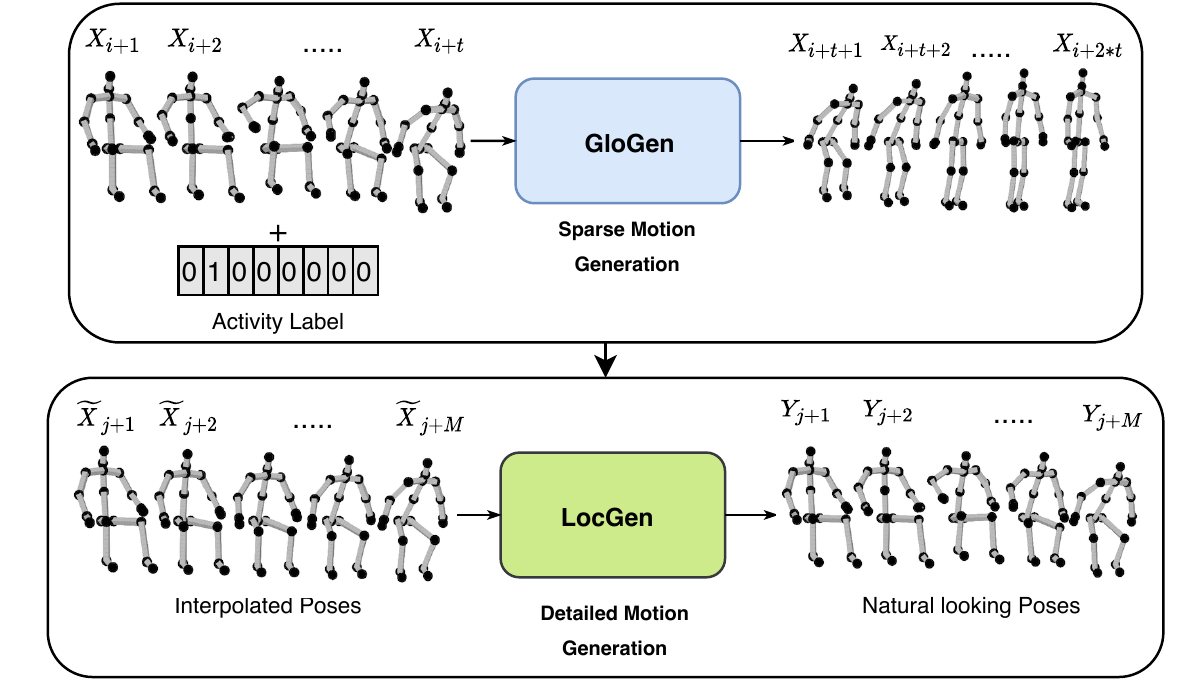}
\end{center}
   \caption{Overview of our two-stage framework, GlocalNet. In the first stage, GloGen generates the sparse motion trajectory of an activity, followed by the second stage, LocGen, that predicts the dense poses from the generated sparse motion.}
\label{fig:our_method}
\end{figure*}
Second, the model only learns the temporally forward dependency on short-term sequences (again with temporally redundant poses) and hence miss to exploit the temporally backward long-term dependencies in poses. 
Third, the majority of these methods do not attempt the conditional generation across a large class of activities. This is probably because there could be a significant amount of partial overlap of short-term pose trajectories across multiple activity classes. Thus, modeling the long-term pose dependency is critical for learning a generalized model.

Recently, graph convolution networks (GCN), that are traditionally used in an action recognition task, are employed to synthesize human motion sequence. GCN based methods~\cite{yan2019convolutional,yu2020structure} model intra-frame (joint level spatial graph) and inter-frame (frame level temporal graph) relations as one spatio-temporal graph for every sequence and perform graph convolution. However, these methods also have multiple limitations that are discussed in detail in Section~\ref{sec:literature}.

This paper aims to overcome the limitations of existing methods and synthesize a long-term human motion trajectory across a large variety of human activity classes ($>50$). We propose a two-stage activity generation method to achieve this goal, where the first stage deals with learning the long-term global pose dependencies in activity sequences by learning to synthesize a sparse motion trajectory while the second stage addresses the generation of dense motion trajectories taking the output of the first stage.
 \begin{figure*}[ht!]
\begin{center}
\includegraphics[width=\linewidth]{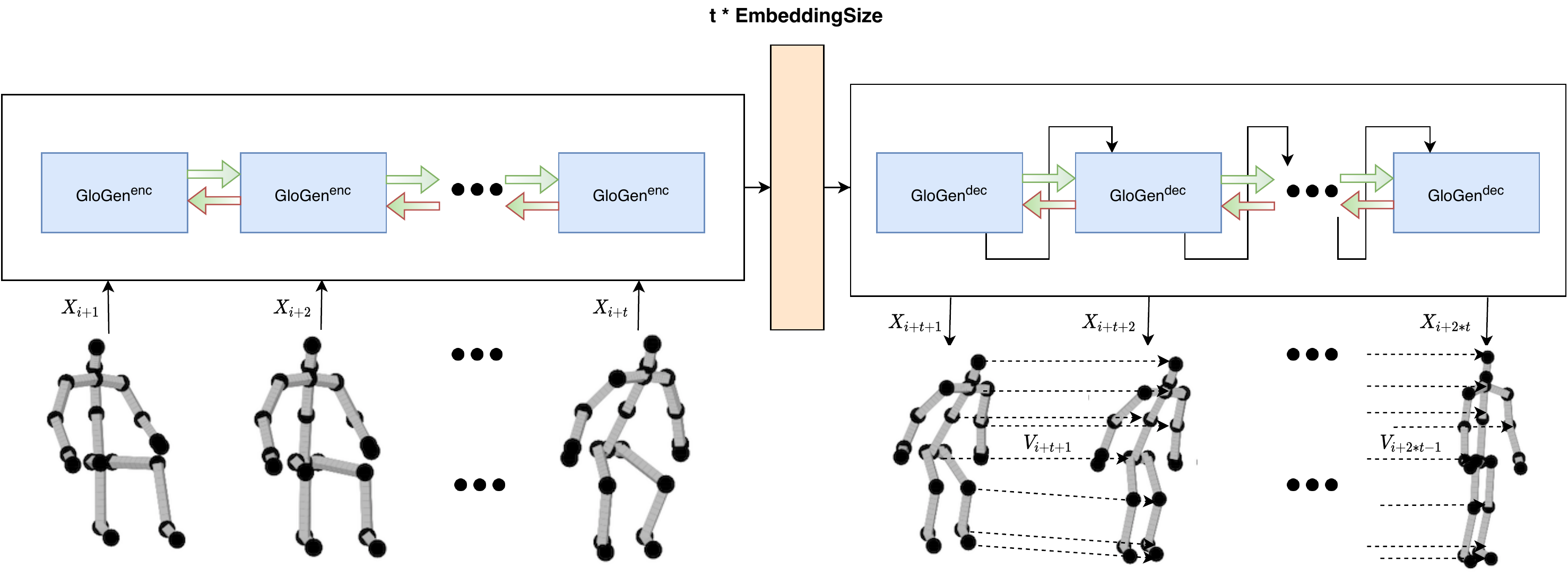}
\end{center}
\caption{Architecture of GloGen network used as sparse motion trajectory generator. }
\label{fig:glogen_architecture}
\end{figure*}

We demonstrate the superiority of the proposed method over SOTA methods using various quantitative evaluation metrics on publicly available datasets \cite{ionescu2013human3, shahroudy2016ntu, cmudataset}, where our method generalizes well even on $60$ activity classes. As shown in Figure~\ref{fig:motivation_fig_initial_pose}, our method is capable of generating the different types of activities based on input class labels and in Figure~\ref{fig:motivation_fig_seq_example} we demonstrate the transition between Standing Up and Drinking activity.
%
Following are the key contributions of our work:
\begin{itemize}
  \item We propose a novel two-stage deep learning method to synthesize long-term ($> 6000$ ms) dense human motion trajectories. 
  \item Our method is capable of generating class-aware motion trajectories. The proposed GloGen embed the sparse activity sequences into a lower dimensional discriminative subspace enabling generalization to a large number of activity classes.
  \item Proposed method can generate a new motion trajectory as a temporal sequence of multiple activity types.
  \item Proposed method can control the pace of generated activities, thereby enabling the generation of variable speed motion trajectories of the same activity type.
  \item To the best of our knowledge, our method first time demonstrates the generalization ability of any long-term ($> 6000$ ms) motion trajectory synthesis method over $60$ activity classes.
\end{itemize}

\section{Related Work}
\label{sec:literature}
Traditional methods~\cite{lee2006human,martinez2013generalized,kovar2008motion,casas20124d} used graph-based modeling of poses for motion trajectory synthesis. 
Majority of the recent deep learning methods aimed at short or medium-term motion synthesis and that limited to a single or small set of activity classes. 
%
%
\cite{holden2016deep} used foot and ground contact information to synthesize locomotion tasks over a given trajectory using a convolutional autoencoder. However, the proposed approach is limited to the locomotion task only and cannot synthesize any other type of action. In \cite{yan2019convolutional}, the authors proposed a method to generate human motion using a graph convolution network.

RNN based approaches have performed well for action recognition, as shown in \cite{li2018independently}. Several researchers followed a similar direction to solve the task of human motion synthesis and proposed approaches based on RNNs. Kundu et al. \cite{kundu2019unsupervised} proposed a method for the task of human motion synthesis using an LSTM autoencoder setup. The proposed network encodes and then decodes back a given motion but is not capable of generating any novel human motion. In \cite{ghosh2017learning}, the authors proposed an approach to generate human motion using the LSTM autoencoder setup. In \cite{habibie2017recurrent} authors proposed a variational autoencoder setup to generate human motion. In \cite{martinez2017human} the network is trained on multiple actions, but they didn't provide any way to control the type of output motion trajectory.


\begin{figure*}[ht!]
\begin{center}
    \includegraphics[width=\linewidth,trim={0 1.2cm 0 0},clip]{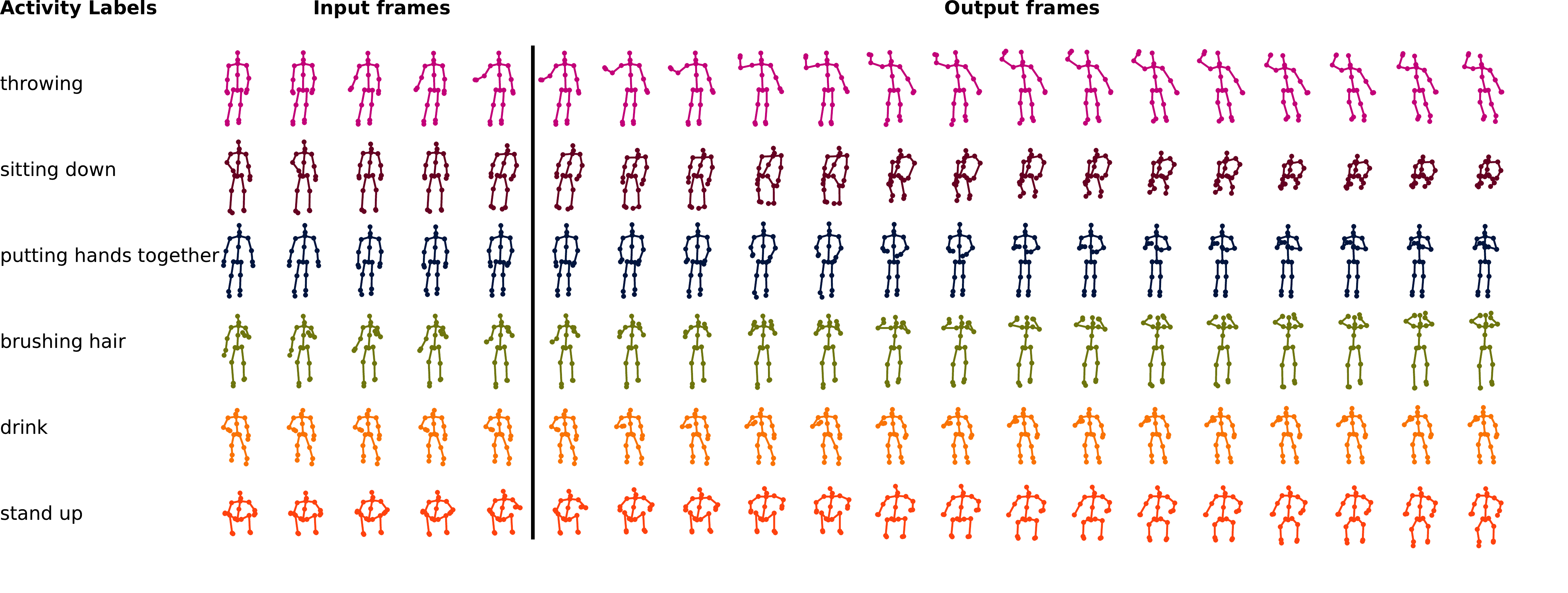}
\end{center}
\caption{Output of GloGen using different activity labels and initial poses. }
\label{fig:multiple_activities}
\end{figure*}
There has been a significant increase in applications and performance of generative models with the arrival of GAN \cite{goodfellow2014generative}. Generative adversarial networks were originally proposed to generate images and later on for videos. Recent methods attempted to synthesize better human motion by incorporating GANs with RNNs in Seq2Seq autoencoders. In \cite{kiasari2018human} Kiasari et al. proposed a method to generate human motion using labels starting poses and a random vector to synthesize human motion, but they did not provide any quantitative results in the paper, and qualitative analysis is also unsatisfactory. In \cite{barsoum2018hp}, the authors proposed an approach to generate human motion using GAN.

Recent GCN-based method~\cite{yu2020structure} models a sequence as a spatio-temporal graph and perform class conditioned graph convolution. However, their fixed size graph modeling limits their scalability to generate long-term sequences. More importantly, the size of the frame sequence that can be considered for learning the temporal dependencies across frames/poses is shown to be relatively small. Additionally, since their method takes random noise as input, it lacks control using the initial state of the activity and hence is not capable of transitioning between two actions as done by our method in Figure \ref{fig:motivation_fig_seq_example}. Similarly, one can not synthesize a long duration motion sequence by repeatedly invoking their fixed length GCN generator. Another similar work in~\cite{yan2019convolutional} proposed to synthesize very long-term sequences but fails to model class conditioning in their generative model, which is an essential aspect of motion synthesis. 


\section{Our Method: GlocalNet}

Our novel two-stage human motion synthesis method attempts to address the key challenges associated with the task of long-term human motion trajectory synthesis across a large number of activity classes.
More precisely, we aim to learn the long-term temporal dependencies among poses, cyclic repetition across poses, bi-directional, and multi-scale dependencies among poses. Additionally, our method attempts to incorporate class priors in the generation process to learn a discriminatory embedding space for motion trajectories, thereby addressing the generalisability aspect over a large class of human activities. \\

\noindent\textbf{Two Stage Motion Synthesis} \\
\label{sec:method_human_motion_synthesis}\\
The key limitation of the existing temporal auto-regressive models like Seq2Seq is the Markovian dependency assumption, where a new set of poses is assumed to be depending upon just a few preceding poses. This impairs their capability to capture the long-term dependence among poses that are far apart and thus led to an accumulation of the prediction error (e.g., mean joint error) while attempting iterative prediction of long-term motion trajectories. We propose to overcome this limitation by splitting the process into two stages, where the first stage is employed to capture the global dependence among poses by learning temporal models on sparsely sampled poses instead of original dense motion trajectories. Thus, the second stage can subsequently deal with the generation of more detailed motion trajectories starting from sparse motion trajectories synthesized by the first stage. This also enables the additional capability to control the frame rate of the synthesized motion trajectories. 

The other key drawback of the Markovian model is its incompetence to exploit the temporally backward dependencies in poses. Thus, we propose to employ the bi-directional LSTMs in the first stage to overcome this limitation. 
Finally, existing methods fail to generalize the motion synthesis for a large class of activity types, probably because of significant overlap among motion trajectories across multiple classes. We propose to overcome this limitation by employing a conditional generator (with class prior) in the first stage itself (while generating sparse global motion trajectories). 

Such decoupling enables the first stage to learn the class-specific long-term (bi-directional) pose dependence while the second stage primarily focuses on the generation of class agnostic fine-grained dense motion trajectories given the sparse output trajectories from the first stage. Figure \ref{fig:our_method} outlines the overview of our proposed two-stage method.
\subsection{First Stage: GloGen}
The first stage is implemented as auto-regressive Seq2Seq network equipped with bi-directional LSTMs called {\emph GloGen}, shown in Figure~\ref{fig:glogen_architecture}. 
The GloGen encoder takes as input a sequence of a sparse set of $t$ initial poses $\{X_{1}, X_{2}... X_{t}\}$ that are uniformly sampled from input motion trajectory during training. Here each pose $X_{i}$ depicts a fixed dimensional vectorial representation of the human pose. These poses are then concatenated with the action class priors encoded as one-hot vectors and fed to the encoder. Unlike traditional Seq2Seq models, we feed all the output states of the encoder i.e., $\{ H_{1}, H_{2}... H_{t}\}$  as input to the GloGen decoder instead of just the last state. The rationale behind this choice is that all hidden states jointly capture the sparse input poses' global embedding. 
Finally, the decoder output is considered as the set of predicted $t$ number of poses. These predicted poses are used as input to synthesize the next set of $t$ iteratively to generate the sparse global motion.
\begin{equation}
\begin{gathered}
H_{i+1}, H_{i+2}... H_{i+t} = \\
GloGen Encoder(X_{i+1}, X_{i+2}... X_{i+t})
\end{gathered}
\end{equation}

\begin{equation}
\begin{gathered}
X_{i+t+1}, X_{i+t+2}... X_{i+2t} = \\
GloGen Decoder(H_{i+1}, H_{i+2}... H_{i+t})
\end{gathered}
\end{equation}

\begin{figure}
\begin{center}
\includegraphics[width=\linewidth]{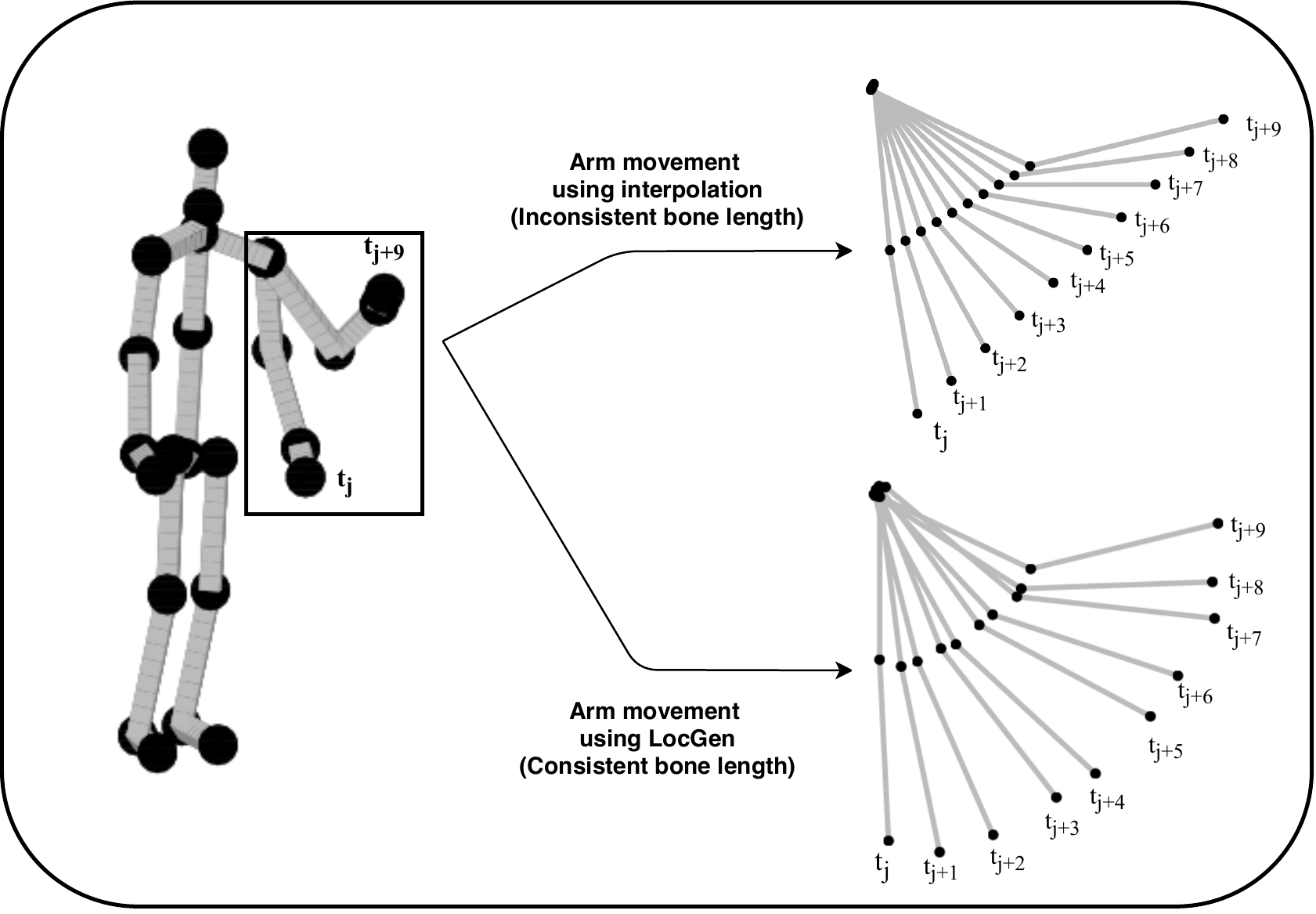}
\end{center}
   \caption{Comparison of linear interpolation v/s LocGen based generation of dense motion trajectories.}
\label{fig:short}
\label{fig:local_gen_arm}
\end{figure}

\begin{table}[b!]
\begin{center}
\begin{tabular}{|l|c|c|}
\hline
Method & LocGen $\downarrow$ & Interpolation $\downarrow$ \\ \hline
Vae Seq2Seq & 0.222 & 0.230 \\
Seq2Seq \cite{martinez2017human} & 0.214 & 0.223 \\
att. Seq2Seq \cite{vaswani2017attention} & 0.336 & 0.352 \\
acLSTM \cite{li2017auto}  & 0.328 & 0.355 \\
\hline
Our Method & \textbf{0.172} & \textbf{0.177} \\ \hline
\end{tabular}
\end{center}
\caption{Comparison of Our Method GlocalNet (GloGen + LocGen) in terms of Euclidean Distance per frame on NTU RGB+D(3D) dataset. LocGen majorly contributes to the qualitative results rather than quantitative.}
\label{tab:LocGen_GloGen_table}
\end{table}

\begin{table*}[t!]
\begin{center}
\begin{tabular}{|l|c|c|c|c|}
 \hline
 \multirow{2}{3cm}{Models} & \multicolumn{2}{c|}{cross-view} & \multicolumn{2}{c|}{cross-subject}\\
 \cline{2-5}
 & MMD$_{avg} $ $\downarrow$ & MMD$_{seq}$ $\downarrow$ & MMD$_{avg}$ $\downarrow$ & MMD$_{seq}$ $\downarrow$ \\
 \hline
 SkeletonVAE \cite{habibie2017recurrent} & 1.079 & 1.205 & 0.992 & 1.136 \\
 SkeletonGAN \cite{cai2018deep} & 0.999 & 1.311 & 0.698 & 0.788 \\
 c-SkeletonGAN \cite{wang2019learning} & 0.371 & 0.398 & 0.338 & 0.402 \\
 SA-GCN \cite{yu2020structure} & 0.316 & 0.335 & 0.285 & 0.299 \\
  \hline
Our Method ($L_{J}$) & 0.213 & 0.218 & 0.201 & 0.212\\
Our Method ($L_{MF}$) & 0.646 & 0.647 & 0.601 & 0.625\\
\textbf{Our Method ($L_{J} + L_{MF}$)} & \textbf{0.195} & \textbf{0.197} & \textbf{0.177} &\textbf{0.187} \\
  \hline
 \end{tabular}
\end{center}
\caption{Comparison of Our Method (GloGen) in terms of MMD on NTU RGB+D(2D).}
\label{tab:GloGen_table_NTU}
\end{table*}

\begin{table}[hb!]
\begin{center}
\begin{tabular}{|l|c|c|}
 \hline
 Models & MMD$_{avg}$ $\downarrow$ & MMD$_{seq}$ $\downarrow$\\
 \hline
 E2E \cite{wichers2018hierarchical}  & 0.991 & 0.805 \\
 EPVA \cite{wichers2018hierarchical} & 0.996 & 0.806 \\
 adv-EPVA \cite{wichers2018hierarchical} & 0.977 & 0.792 \\
 SkeletonVAE \cite{habibie2017recurrent}   & 0.452 & 0.467 \\
 SkeletonGAN \cite{cai2018deep}  & 0.419 & 0.436 \\
 c-SkeletonGAN \cite{wang2019learning} & 0.195 & 0.218 \\
 SA-GCN \cite{yu2020structure} & 0.146 & 0.134 \\
 \hline
 Our Method  & \textbf{0.103} & \textbf{0.102} \\
 \hline
 \end{tabular}
\end{center}
\caption{Comparison of Our Method (GloGen) in terms of MMD on Human 3.6M.}
\label{tab:GloGen_table_h36m}
\end{table}

\subsection{Second Stage: LocGen}
Once we predict the sparse motion trajectories from GloGen, we need to process them further to obtain dense motion trajectories as the predicted pose will be far apart in pose space and hence would lack the temporal smoothness behavior. One option to obtain a dense set of poses from sparse-poses is to perform simple interpolation based upsampling in Euclidean representation of poses. However, from Figure~\ref{fig:local_gen_arm}, we can infer that simple interpolation is not a good option as it leads to unnatural motion trajectories. This is because the intermediate poses provided by the interpolation typically yield straight lines due to which fix bone length constraint is violated frequently, and the motion does not seem natural. Interpolation in Euler angle space is an alternate option that do not violate bone-length constraint. However, such representation of skeleton has issue 
that even small error in joint angles near root of kinematic tree results in large error in the joint locations for other dependent joints, while doing interpolation. Thus, we stick to Euclidean $[x,y,z]$ representation of joints in this work but other representations can also be considered. 

We propose to obtain dense motion trajectories using another auto-regressive network named {\emph LocGen}, shown in Figure~\ref{fig:our_method}. Input to LocGen encoder is a set of (Euclidean) interpolated poses. The encoder first embeds the human pose into a higher dimension and then fed the hidden states to the decoder (similar to GloGen), generating more natural motion trajectories. LocGen has the same architecture as GloGen except that instead of sparse motion poses, LocGen takes interpolated dense motion trajectories as input, and there is no class prior concatenated with input poses. Thus, LocGen learns to transform interpolated trajectories into natural looking temporally smooth motion trajectories. 

In order to generate interpolated poses between given two sparse-poses generated by GloGen, we use the following formulation. Let $M$ be the number of interpolated poses that need to be synthesized between two given sparse-poses $X_{i}$ and $X_{i+1}$. Let $\widetilde{X}_{j}$ be the $j$-th interpolated pose for $ 1\leq j  \leq M $, then we can compute $\widetilde{X}_{j}$ as:
\begin{equation}
\label{eq:lin-interpolate}
\widetilde{X}_{j} = \alpha_{j}*X_{i} + (1-\alpha_{j})*X_{i+1}
\end{equation}
where $\alpha_{j} = j/M$.\\
\\
\{$\widetilde{X}_{j+1}$, $\widetilde{X}_{j+2}$ ... $\widetilde{X}_{j+M}$\} are given as input to the LocGen which first embeds them into the higher dimension and then use the embeddings to generate natural looking poses\{$Y_{j+1}$, $Y_{j+2}$ ... $Y_{j+M}$\}.

\begin{equation}
Y_{j+1}, Y_{j+2}... Y_{j+M} = LocGen(\widetilde{X}_{j+1}, \widetilde{X}_{j+2}... \widetilde{X}_{j+M})
\end{equation}




\section{Experiments \& Results}

Every model is trained individually from scratch using same setting in Table~\ref{tab:LocGen_GloGen_table}. All of the trained models, code, and data shall be made publicly available, along with a working demo. Please refer to our supplementary material for an extended set of video results.

\subsection{Datasets}

\noindent \textbf{Human 3.6M \cite{ionescu2013human3}:} Following the same pre-processing procedure as in \cite{wang2019learning}, we down-sampled 50 Hz video frames to 16 Hz to obtain better representative and larger variation 2D human motions. The skeletons consist of 15 major body joints, which are represented in 2D. We consider ten distinctive classes of actions in our experiments, that includes sitting down, walking, direction, discussion, sitting, phoning, eating, posing, greeting, and smoking. \\

\noindent \textbf{NTU RGB+D(3D)~\cite{shahroudy2016ntu}:} This dataset contains around 56,000 samples on 60 classes performed by 40 subjects and recorded with 3 different cameras. Hence, it provides a good benchmark to test 3D human motion synthesis. We have used the available Cross-Subject split provided by the dataset for our experiments. We resort to standard pre-processing steps adopted by existing methods~\cite{kundu2019unsupervised}. \\

\noindent \textbf{NTU RGB+D(2D)~\cite{shahroudy2016ntu}:}
 To compare with previous works \cite{yu2020structure}, we follow the same setting to obtain 2D coordinates of 25 body joints and consider the same ten classes to run experiments. We use the available Cross-View and Cross-Subject splits. \\
 
\noindent \textbf{CMU Dataset~\cite{cmudataset}:} The dataset is given as sequences of the 3D skeleton with $57$ joints. We evaluate our method on three distinct classes from the CMU motion capture database, namely, martial arts, Indian dance, and walking similar to~\cite{li2017auto}.

\subsection{Implementation Details }

\noindent \textbf{Network Training:} We use Nvidia's GTX 1080Ti, with 11GB of VRAM to train our models. For training GLoGen, the output dimension of our Encoder is 200. We are using 1 layered Bi-LSTM as our Encoder as well as Decoder. Dropout regularization with a 0.25 discard probability, was used for the layers. We use the AdamW optimizer \cite{loshchilov2017decoupled} with an initial learning rate of 0.002, to get optimal performance on our setup. We use MSE loss to calculate our objective function. Similar to \cite{yu2020structure}, we set the predicted action sequence length for Human 3.6M and NTU RGB+D(2D) datasets to be 50 and input sequence length to be 10. We set the batch size for training to be 100, for testing to be 1000. For datasets CMU and NTU RGB+D(3D), a batch size of 64 is used. For training on NTU RGB+D(3D) with all 60 classes, we use input action sequence length to be 5 and predicted sequence length of sparse poses to be 15 for GloGen and then using LocGen, we generate 4 new poses for every pair of adjacent sparse-poses.\\

\begin{table*}[t!]
\begin{center}
\begin{tabular}{|l| c c c c c c c c|}
 \hline
 Method        & 80ms & 160ms & 240ms & 320ms & 400ms & 480ms & 560ms & 640ms\\
 \hline
 \textbf{Walking} & & & & & & & &\\
\hline
acLSTM \cite{li2017auto} & 1.05 & 1.77 & 2.20 & 2.46 & 2.66 & 2.79 & 2.99 & 3.24 \\
 \hline

Scheduled Sampling  \cite{bengio2015scheduled}  & 0.42 & 0.56 & 0.71 & 0.83 & 0.93 & 0.99 & 1.02 & 1.05 \\
 \hline
Seq2Seq \cite{martinez2017human} & \textbf{0.09} & \textbf{0.13} & \textbf{0.24} & \textbf{0.42} & 0.74 & 1.22 & 1.85 & 2.79\\
 \hline
Our Method & 0.36 & 0.47 & 0.52 & 0.60 & \textbf{0.62} & \textbf{0.65} & \textbf{0.71} & \textbf{0.82} \\
 \hline

 \hline
 \textbf{Indian Dance} & & & & & & & &\\
\hline
acLSTM \cite{li2017auto}      & 0.685 & 0.99 & 1.22 & 1.53 & 1.89 & 2.08 & 2.27 & 2.55 \\
 \hline
Scheduled Sampling  \cite{bengio2015scheduled}  &  1.54 & 2.24 & 2.49 & 2.52 & 2.65 & 2.90 & 2.94 & 3.12 \\
 \hline
Seq2Seq \cite{martinez2017human} & \textbf{0.49} & 0.79 & 1.48 & 2.95 & 5.41 & 8.88 & 13.29 & 18.73\\
 \hline
Our Method       & 0.50 & \textbf{0.56} & \textbf{0.64} & \textbf{0.68} & \textbf{0.69} & \textbf{0.69} & \textbf{0.79} & \textbf{0.84}\\
 \hline

 \hline
 \textbf{Martial Arts} & & & & & & & &\\
\hline
acLSTM \cite{li2017auto}       & 0.52 & 0.74 & 0.95 & 1.14 & 1.35 & 1.56 & 1.73 & 1.88 \\
 \hline

Scheduled Sampling  \cite{bengio2015scheduled}  &  0.63 & 0.86 & 0.91 & 0.98 & 1.07 & 1.12 & 1.20 & 1.28 \\
 \hline
Seq2Seq \cite{martinez2017human} & \textbf{0.28} & \textbf{0.43} & 0.87 & 1.57 & 2.53 & 3.89 & 5.83 & 8.62\\
 \hline
Our Method        & 0.40 & \textbf{0.43} & \textbf{0.47} & \textbf{0.52} & \textbf{0.55} & \textbf{0.59} & \textbf{0.67} & \textbf{0.71}\\
 \hline

\end{tabular}
\end{center}
\caption{Comparison of Our Method (GloGen) in terms of Euclidean Distance per frame on CMU dataset.}

\label{tab:GloGen_table_cmu}
\end{table*}


\noindent \textbf{Loss Function:} Loss function is calculated on joint locations and motion flow. 
We use the following loss function to train out network $L_{J}$ and $L_{MF}$.
\begin{equation}
    L = (\lambda_{1} * L_{J}) + (\lambda_{2} * L_{MF})
    \label{eqn:our_method_loss_function}
\end{equation}
The joint loss $L_{J}$ in Equation~\ref{eqn:joint_loss} gives the vertex-wise Euclidean distance between the predicted joints $X_{i}$ and ground truth joints $\hat{X_{i+1}}$.
\begin{equation}
\label{eqn:joint_loss}
    L_{J} = \sum_{i=1}^{t}||X[i] - \hat{X[i]}||_{2}
\end{equation}
In order to enforce smoothness in temporal sequence, we minimize the motion flow loss $L_{MF}$ defined in Equation~\ref{eqn:mf_loss}, which gives the Euclidean distance between the predicted motion flow $V_{i}$ and ground truth motion flow $\hat{V_{i+1}}$.
\begin{equation}
\label{eqn:mf_loss}
    L_{MF} = \sum_{i=1}^{t-1}||V[i] - \hat{V[i]}||_{2}
\end{equation}
Where, motion flow for the i\textsuperscript{th} frame $\hat{V_{i+1}}$. is the difference between joint locations $\hat{X_{i+1}}$ and $\hat{X_{i}}$.
\begin{equation}
    \hat{V_{i}} = \hat{X_{i+1}} - \hat{X_{i}}
\end{equation}

\subsection{Evaluation Metrics}

\noindent \textbf{Maximum Mean Discrepancy:} 
The metric is based on a two-sample test to measure the discrepancy of two distributions based on their samples. The metric has been used in \cite{walker2017pose, wang2019learning, yu2020structure} for measuring the quality of action sequences by evaluating the similarity between generated actions and the ground truth. Similar to \cite{wang2019learning}, for calculating MMD on motion dynamics which are in the form of sequential data points, the average MMD over each frame is denoted by MMD$_{avg}$ and MMD over whole sequences are denoted by MMD$_{seq}$. \\

\noindent \textbf{Euclidean distance:} This metric used in \cite{li2017auto} calculates error as the euclidean distance from the ground truth for the corresponding frame.

\subsection{Results}


\noindent\textbf{Long-term Dense Motion Synthesis: }
We use GlocalNet to generate long-term dense motion sequences. Table~\ref{tab:LocGen_GloGen_table} shows the results on NTU RGB+D(3D) for dense motion trajectory synthesis and compare it with existing methods. All the methods were trained from scratch using the same data pre-processing~\cite{kundu2019unsupervised} and have the same input(Class Label \& Initial Poses). These quantitative results show the superior performance of the GlocalNet. Additionally, we report detailed results including long term motion ($> 6000$ ms) and class-wise performance in the supplementary material. We can clearly infer that our proposed solution outperforms all the existing methods.
\begin{figure*}[t!]
\begin{center}
    \includegraphics[width=0.9\linewidth]{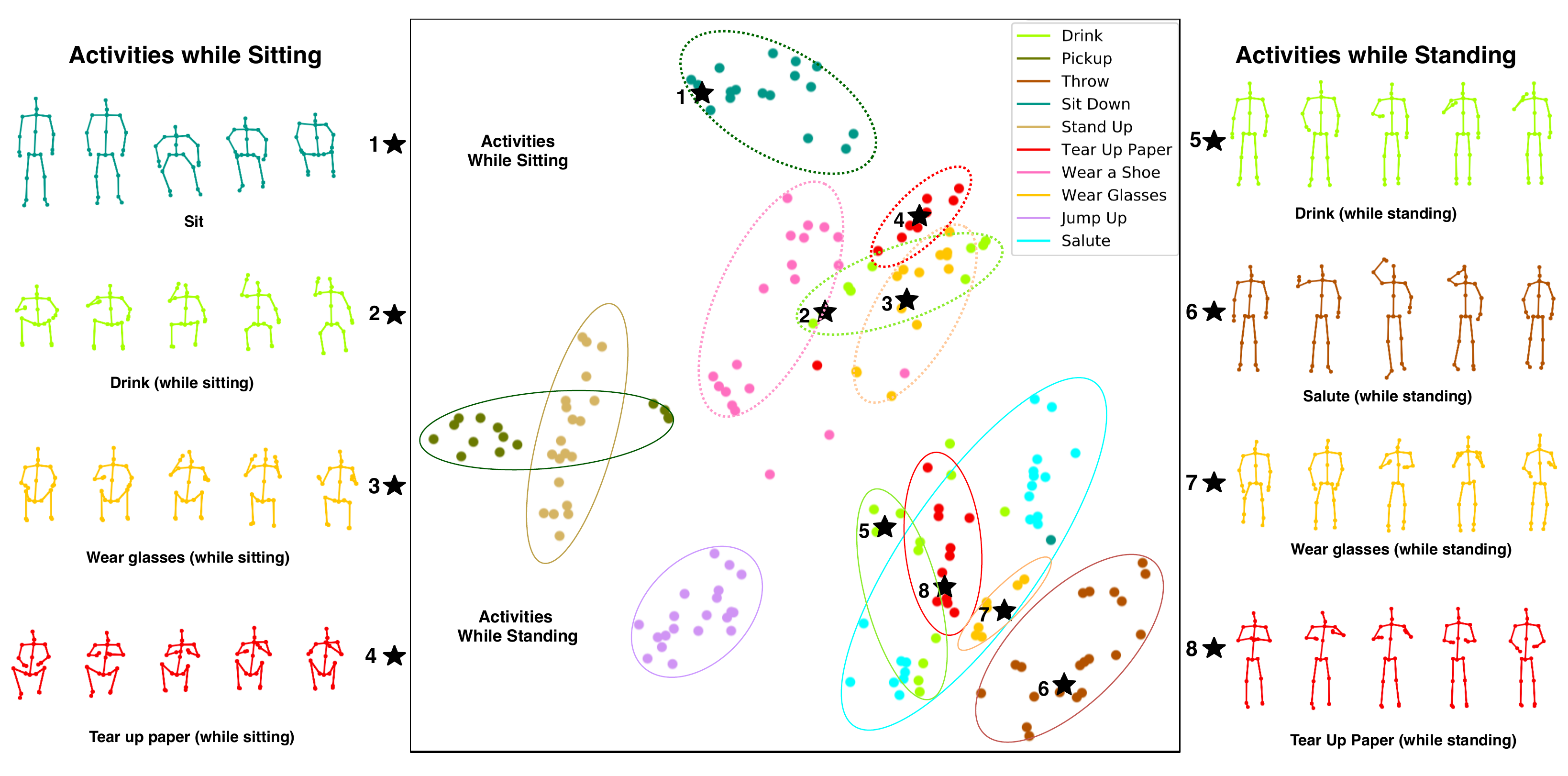}
\end{center}
\caption{The t-SNE plot of GloGen embedding subspace along with the plot of selected motion trajectories where multiple samples for different classes are represented as color-coded 3D points. 
}
\label{fig:embeddings}
\end{figure*}
Figure~\ref{fig:multiple_activities} depicts the synthesized sparse motion trajectories obtained using the GlocalNet on NTU RGB+D(3D) dataset for six different activity classes. As we can see from the figure, the network is able to learn the global long-term temporal dependence in poses successfully across multiple classes and thus generate significantly different motion trajectories for similar initial input poses.\\
\\
\noindent
\textbf{Comparison with Short-term Motion Synthesis Models: }
To compare with existing short-term motion synthesis models on different datasets, we use the first stage of our network(GloGen). For fair comparison, we follow the same settings as followed in these methods. Table~\ref{tab:GloGen_table_NTU} contains the quantitative results on NTU RGB+D(2D) and our method outperforms others with a good margin. Table~\ref{tab:GloGen_table_h36m} shows the results on Human 3.6M for GloGen, which outputs sparse-motion trajectory and compare with SOTA methods. These quantitative results suggest the superior performance of the GloGen over the MMD metric. Additionally, as shown in Table~\ref{tab:GloGen_table_cmu} for CMU Dataset, we report superior performance of our method over the existing ones on Euclidean per frame metric. As reported in the table, our method shows consistent performance even for longer sequences across different actions.

\noindent \textbf{Ablation Study on Loss Functions:} In order to show the importance of the proposed $L_{J}$ and $L_{MF}$ loss separately, we also trained our network using the individual loss components and reported the results in Table \ref{tab:GloGen_table_NTU}. As it is clearly visible, $L_{MF}$ alone is not sufficient; in combination with $L_{J}$ it helps improve the performance of our method. In terms of qualitative results, we observed jitters in the generated sequence (without having $L_{MF}$). Thus, $L_{MF}$ enables the network to learn generating smoother transition in skeleton sequences.  

\noindent \textbf{Synthesis for Sequence of Activities:} Our network can also be used to generate a multi-activity motion trajectory by temporally varying the activity prior. To achieve this, we first synthesize the motion trajectories using the approach described in Section \ref{sec:method_human_motion_synthesis}. Then we treat the final $t$ poses of the generated trajectory as the initial $t$ poses for generating the next set of $t$ poses belonging to new action class by providing the one-hot vector for the new class prior. This process is repeated to generate a new sequence with potentially multiple activity classes, in a single synthesized sequence of arbitrary length.

Figure~\ref{fig:motivation_fig_seq_example} shows an example of a sparse motion trajectory where we generate poses for Stand Up activity and then use its last set of poses to generate Drink activity. Here, we can clearly visualize a smooth transition of poses across the two classes of activities.



\section{Discussion}
\label{sec:Discussion}
\noindent

A major limitation of the Seq2Seq models class is that the last encoder hidden state becomes the bottleneck of the network as all the information at the input side passes through it to reach the decoder. To deal with this problem, attention architecture was proposed  \cite{vaswani2017attention}, where all the encoder hidden states are given to the decoder along with affinity scores that tell the importance of every input state corresponding to every output state. Such attention enabled Seq2Seq networks to achieve SOTA performance for the task of machine translation. However,  generating motion is a different task from machine translation as we aim to predict the future poses looking at the previous ones, while modeling the long-term global dependency in far away poses. Therefore, in our method, instead of giving only the last state, we share the outputs of all states from the encoder to decoder LSTM units and predict the future poses.\newline


\noindent \textbf{GloGen Embedding Subspace: }
In order to visualize the behavior of feature embeddings, we concatenate the pose embeddings of GloGen-encoder over a sequence and project it as a point into 2D space using t-SNE. Figure \ref{fig:embeddings} shows the t-SNE plot of embedding subspace along with the skeleton representation of selected motion trajectories where multiple samples for different classes are represented as color-coded 2D points. We can clearly infer from this figure that proposed GloGen projects these sequences into a discriminative subspace that enables it to handle the synthesis of different classes better. 
Interestingly, we can also see that some sequences from a few activities are scattered across two clusters as they can be performed while both sitting or standing, e.g., Wear glasses and Drink. Nevertheless, apart from a few outlier points due to the noisy samples present in the NTU RGB+D(3D) dataset, this plot clearly indicates the subspace's class discriminative nature.

\section{Conclusion}

 In this paper, we propose a novel two-stage method for synthesizing long-term human-motion trajectories across a large variety of activity types. The proposed method can also generate new motion trajectories as a combination of multiple activity types as well as allows us to control the pace of generated activities. We demonstrate the superiority of the proposed method over SOTA methods using various quantitative evaluation metrics on publicly available datasets.

{\small
\bibliographystyle{ieee_fullname}
\bibliography{egbib}
}

\end{document}


%
%

\twocolumn[ \section*{Supplementary Material of "GlocalNet: Class-aware Long-term Human Motion Synthesis"}]


\newline
\subsection*{Performance on Individual Classes}
Across all the classes in the dataset NTU RGB+D(3D) the average Euclidean distance across all the classes is 0.17 and the Standard Deviation is 0.06. We provide the results of top-5 and bottom-5 performing classes in Table \ref{tab:top_5} and \ref{tab:bottom_5}, respectively.

Although the reconstruction error on bottom-5 classes was high, we observed generated sequences were quite reasonable in terms of qualitative visualization but definitely not coherent with respect to Ground Truth.
\begin{table}[h!]
\begin{center}
\begin{tabular}{|p{3.2 cm}|p{3.2 cm}|}
\hline
\textbf{Class} & \textbf{Euclidean Distance} \\ \hline
Playing with Phone & 0.079 \\
Shake Head & 0.080 \\
Typing On Keyboard & 0.093 \\
Put the Palms together & 0.095 \\
Drop & 0.106 \\\hline
\end{tabular}
\end{center}
\caption{Performance of top-5 classes on NTU RGB+D(3D) in terms on Euclidean Loss.}
\label{tab:top_5}
\end{table}

\begin{table}[h!]
\begin{center}
\begin{tabular}{|p{3.2 cm}|p{3.2 cm}|}
\hline
\textbf{Class} & \textbf{Euclidean Distance} \\ \hline
Wear Jacket & 0.345 \\
Throw & 0.292 \\
Hugging & 0.279 \\
Nausea & 0.263 \\
Punching & 0.263 \\\hline
\end{tabular}
\end{center}
\caption{Performance of bottom-5 classes on NTU RGB+D(3D) in terms on Euclidean Loss.}
\label{tab:bottom_5}
\end{table}

\subsection*{Performance without Class Prior}
\begin{figure*}[t!]
\begin{center}
    \includegraphics[width=0.8\linewidth]{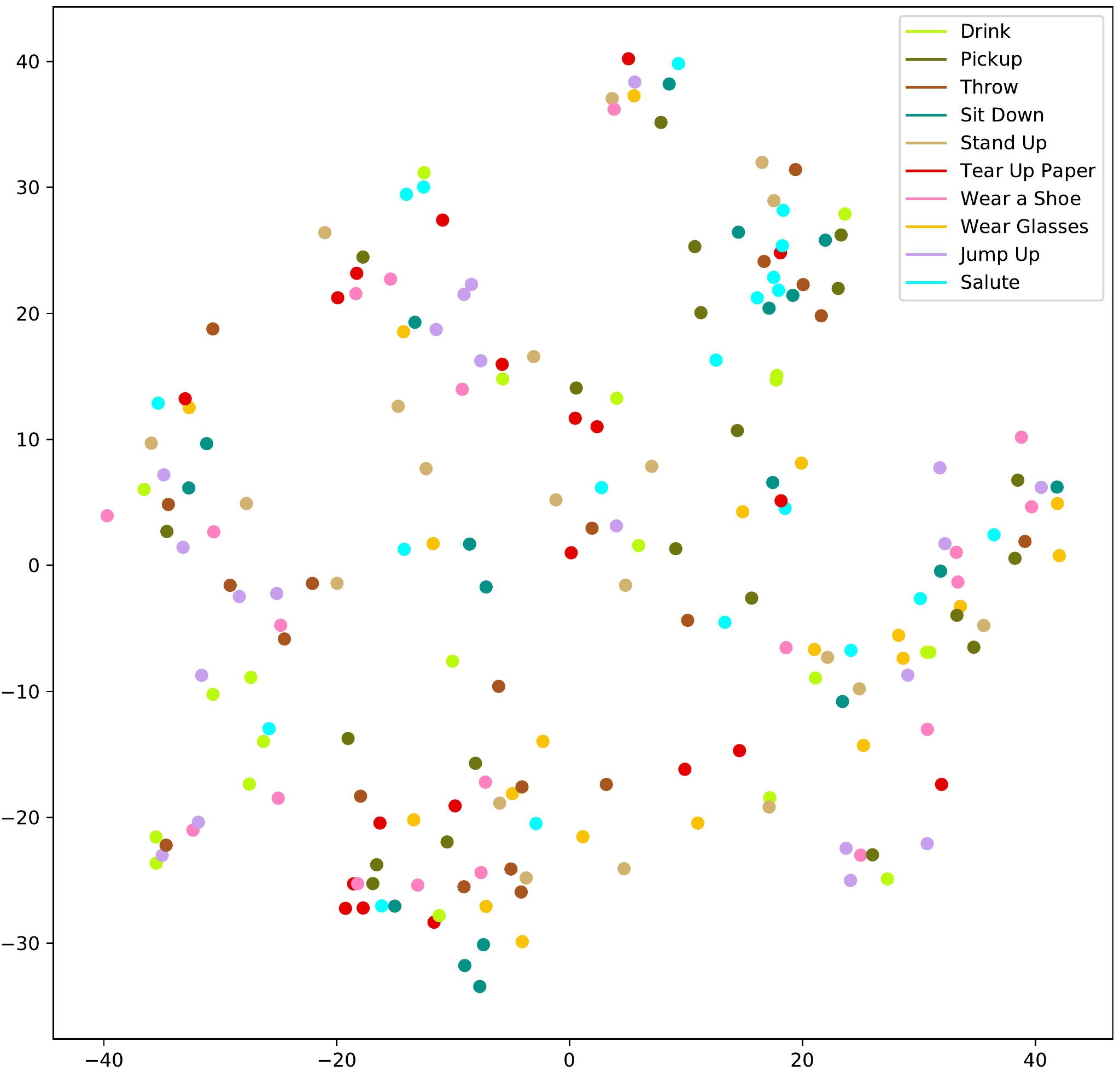}
    
\end{center}
\caption{The t-SNE plot of embedding subspace without using the input class label, here samples for different classes are represented as color-coded 3D points. 
}
\label{fig:embedding_no_class}
\end{figure*}

Our Method shows better performance than previous SOTA models even without the supervision of class prior. However, as visible in Figure \ref{fig:embedding_no_class} the subspace is significantly cluttered when we don't use the class prior as compared to the embedding diagram from main paper (refer Figure 6).


\begin{table*}[h!]
\begin{center}
\begin{tabular}{|l|c|c|c|c|}
 \hline
 \multirow{2}{3cm}{Models} & \multicolumn{2}{c|}{cross-view} & \multicolumn{2}{c|}{cross-subject}\\
 \cline{2-5}
 & MMD$_{avg} $ \downarrow & MMD$_{seq}$ \downarrow & MMD$_{avg}$ \downarrow & MMD$_{seq}$ \downarrow \\
 \hline
 Without Class-prior & 0.226 & 0.231 & 0.186 & 0.193\\
  \hline
 Our Method & \textbf{0.195} & \textbf{0.197} & \textbf{0.177} &\textbf{0.187} \\
 \hline
 \end{tabular}
\end{center}
\caption{Performance without Class label in terms of MMD on NTU RGB+D(2D).}
\label{tab:ablation_study}
\end{table*}

